\documentclass{article}
\usepackage{spconf,amsmath,graphicx}
\usepackage{amsmath}
\usepackage{graphicx}
\usepackage{fancyhdr}
\usepackage{listings} 
\usepackage{changepage} 
\usepackage[figurename=Fig., tablename=Tab., small]{caption}[2008/04/01]
\usepackage{amssymb}
\usepackage{amsthm}
\usepackage{float}
\usepackage{subfigure}

\newcommand*\samethanks[1][\value{footnote}]{\footnotemark[#1]}

\title{Using Deep Cross Modal Hashing and Error Correcting Codes for Improving the Efficiency of Attribute Guided Facial Image Retrieval}

    \name{Veeru Talreja\sthanks{Authors Contributed Equally}, Fariborz Taherkhani\samethanks[1], Matthew C. Valenti , and Nasser M. Nasrabadi }\address{West Virginia University, Morgantown, USA}

\begin{document}
%
\maketitle
\begin{abstract}
With benefits of fast query speed and low storage cost, hashing-based image retrieval approaches have garnered considerable attention from the research community. In this paper, we propose a novel Error-Corrected Deep Cross Modal Hashing (CMH-ECC) method which uses a bitmap specifying the presence of certain facial attributes as an input query to retrieve relevant face images from the database. In this architecture, we generate compact hash codes using an end-to-end deep learning module, which effectively captures the inherent relationships between the face and attribute modality. We also integrate our deep learning module with forward error correction codes to further reduce the distance between different modalities of the same subject. Specifically, the properties of deep hashing and forward error correction codes are exploited to design a cross modal hashing framework with high retrieval performance. Experimental results using two standard datasets with facial attributes-image modalities indicate that our CMH-ECC face image retrieval model outperforms most of the current  attribute-based face image retrieval approaches.
\end{abstract}
\section{Introduction}
\label{sec:intro}

With the fast development of search engines and social networks, there exists a vast amount of  multimedia data, such as texts, images and videos being generated on the world wide web everyday. The presence of multimedia big data has sparked a rise of content based image retrieval (CBIR) techniques in the research community. Approximate nearest neighbors (ANN) based semantic search has garnered a lot of attention to guarantee the retrieval quality and computing efficiency for CBIR in large-scale datasets. Cross-modal retrieval is an important paradigm of CBIR, which works with multimodal data and supports similarity retrieval across different modalities, e.g., retrieval of relevant facial images in response to attribute query such as ``an old woman wearing glasses". In this paper, we address the problem of cross-modal retrieval of relevant face images in response to facial attributes queries by utilizing a deep cross-modal hashing framework in combination with error correcting codes. 


A fast and promising solution to ANN search for cross-modal retrieval is cross-modal hashing (CMH), which compresses high-dimensional data into compact binary codes and maintains the semantic similarity by mapping images of similar content to similar binary codes. CMH returns relevant results of one modality in response to query of another modality, where respective hash codes in the same latent Hamming space are generated for each individual modality. Recently, application of deep learning to hash methods for uni-modal image retrieval \cite{deep_hashing_liu_2016,cao_2017_hashnet} and cross-modal retrieval \cite{Yang_2017_Pairwise,jiang2017deep} have shown that end-to-end learning of feature extraction and hash coding using deep neural networks is more efficient than using the hand-crafted features \cite{Zhang_2014_LSM,Zhen_NIPS2012_4793}. Particularly, it proves beneficial to jointly learn semantic similarity preserving features and also curb the quantization error of binarizing continuous representation to hash codes. 

 Searching for facial images of people including identification in response to a facial attribute query has been investigated in the past \cite{kumar2008facetracer, 2009_vaquero_attribute,2011_siddiquie_image_ranking,taherkhani2018deep}. However, all of these methods use hand-crafted features to perform a cross-modal retrieval. We present a novel CMH framework called CMH-ECC for error-corrected attribute guided deep cross-modal hashing for face-image retrieval from large datasets. The main contributions of this paper include: (1) \textbf{Error-corrected attribute guided deep cross modal hashing (hereon known as CMH-ECC)} : We have designed a novel architecture using deep cross modal hashing for face image retrieval in response to an attribute query. (2) \textbf{Error correcting codes}: We have integrated the deep cross modal hashing with error correcting codes to further reduce the Hamming distance between different modalities of same subject and improve the retrieval efficiency obtained from performing only deep cross modal hashing. (3) \textbf{Scalable cross-modal hash}: Our architecture CMH-ECC performs facial image retrieval using point wise data without requiring pairs or triplets of training inputs, which makes CMH-ECC scalable to large scale datasets.

\begin{figure*}[t]
\centering
\includegraphics[width=18cm, height=6.5cm]{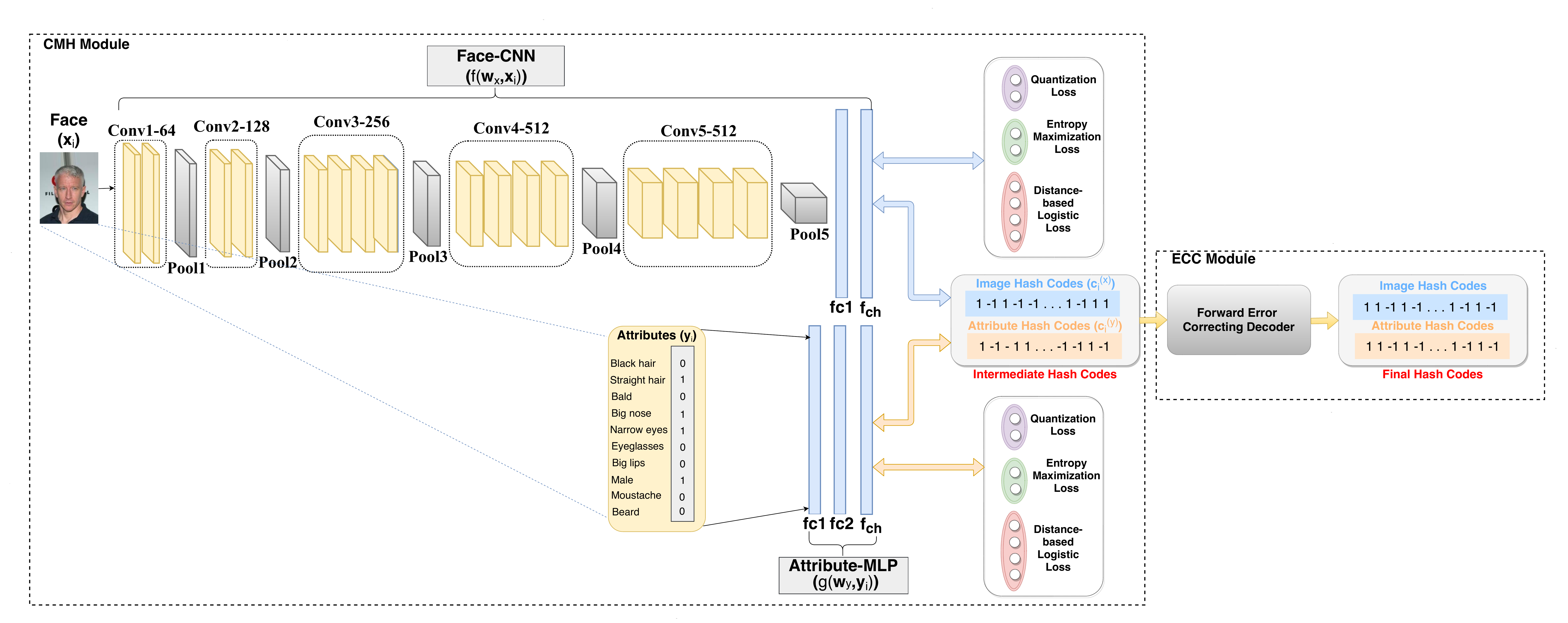}
\vspace{-0.55cm}
\caption{Block Diagram of the CMH-ECC.}\label{fig:arch}
\vspace{-0.45cm}
\end{figure*}

\section{THE PROPOSED CMH-ECC FRAMEWORK}
\label{sec:format}
\subsection{Problem Definition}\label{ssec:probdef}

Define $\mathcal{O}= \{\textbf{o}\textsubscript{i}\}_{i=1}^{n}$ to be the training set where $n$ is the number of  training samples. All the samples have two modalities  $\textbf{X} = \{\textbf{x}\textsubscript{i}\}_{i=1}^{n}$ and $\textbf{Y} = \{\textbf{y}\textsubscript{i}\}_{i=1}^{n}$, which corresponds to image and attribute modalities, respectively. \textbf{x}\textsubscript{i} is the raw image $i$ in a training set of size $n$ and \textbf{y}\textsubscript{i} is  the annotated facial attributes vector related to image $i$. $\textbf{S}$ is a cross-modal similarity matrix  in which $S_{ij} = 1 $ if image $\textbf{x}\textsubscript{i}$ contains a $y_j$ facial attribute, and $S_{ij} = 0$ otherwise. 

Based on the given training information (i.e., \textbf{X}, \textbf{Y} and \textbf{S}),  the proposed method learns two modality-specific hashing functions: $h^{(x)} (\textbf{x}) \in \{-1, +1\} ^ d $ for image modality and $h^{(y)} (\textbf{y}) \in \{-1, +1\} ^ d $ for attribute modality where $d$ is the number of the bits used in the intermediate hash codes. The two hashing functions have to preserve the cross-modal similarity in \textbf{S}. Specifically, if $S_{ij} = 1$, the Hamming distance
between the binary codes $ \textbf{c}^{(x)}_{i}= h^{(x)} (\textbf{x}\textsubscript{i}) $ and $\textbf{c}^{(y)}_{j}= h^{(y)} (\textbf{y}\textsubscript{j}) $ should be small and if $S_{ij} = 0$, the corresponding Hamming distance should be large. The learned hash functions can be employed to generate $d$-bit intermediate hash codes for query and database instances in both modalities. The intermediate hash codes for query and database points are passed through a forward error correcting (FEC) decoder $f^{(d)} (.)$ to generate the final $c$-bit codewords (final hash codes) which are used in the retrieval process.  



The block diagram of the proposed framework is given in Fig. \ref{fig:arch}. The proposed CMH-ECC framework has two modules. The first module is the deep cross modal hashing  module (CMH module) and the second module in the CMH-ECC is the error correcting code module (ECC module).
\vspace{-0.30cm}

\subsection{Deep cross-modal hashing module (CMH)}

 CMH module trains a coupled deep neural network (DNN) to generate intermediate hash codes using a distance-based logistic loss to preserve the cross-modal similarity. The CMH module has three main functions: 1) Learn a coupled DNN using distance-based logistic loss to preserve the cross-modal similarity. 2) In order to preserve a high retrieval performance, control the quantization error for each modality due to the binarization of continuous output activations of the network to hash codes. 3) Maximize the entropy corresponding to each bit to obtain the maximum information provided by the hash codes.

 The CMH module is composed of two networks : A Convolutional Neural Network (CNN) to extract features for image modality and a Multi-Layer Perceptron (MLP) to extract features for facial attribute modality. For CNN network, we have used  VGG-19 \cite{simonyan_very_deep_2014} network pre-trained on the ImageNet \cite{deng_imagenet_2009} dataset as a starting point and fine-tuned it as a classifier by using the CASIA-Web Face dataset. The original VGG-19 consists of five convolutional layers ($conv1-conv5$) and three fully-connected layers ($fc6-fc8$). We discard the $fc8$ layer and replace the $fc7$ layer with a new $f_{ch}$ layer with $d$ hidden nodes, where d is the required intermediate hash code length (the intermediate code length in all the experiments is set to 256 bits). The MLP network comprises three fully connected layers to represent features for the facial attribute modality. To learn attribute features from this network, we annotate each training sample image with a binary bit map that indicates the presence or absence of corresponding facial attribute. This bitmap serves as a facial attribute vector and is used as  input to the MLP network. The first and second layers in the MLP network contain 4,096 nodes with ReLU activation and the number of nodes in the  last fully connected layer is equal to the intermediate hash code length $d$ with identity activation. We use the Adam optimizer \cite{kingma2014adam} with the default hyper-parameter values ($\epsilon = 10^{-3}$, $\beta_1 = 0.9$, $\beta_2 = 0.999$) to train all the parameters using alternative minimization approach. The batch size in all the experiments is fixed to 128. 
 

\begin{figure}[t]
\centering
\includegraphics [scale=0.3]{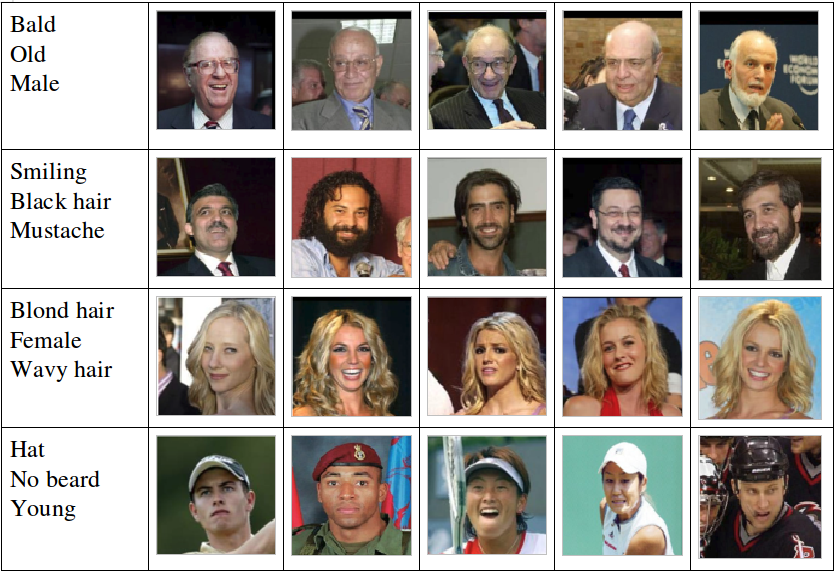}
\caption{Qualitative results: Retrieved images using CMH-ECC for given facial attributes. }\label{fig:results} 
\vspace{-0.45cm}
\end{figure}

 For efficient retrieval results, assuming that two samples $o_i$ and $o_j$ are semantically similar, their corresponding hash codes should also be similar in the low dimensional Hamming space. We design the objective function for generating efficient hash codes. Our objective function for CMH comprises of three parts: (1) distance-based logistic loss; (2) quantization loss; and (3) entropy maximization loss.
 
 \begin{figure*}[t]
\centering
\subfigure[Single Attribute Queries]{\includegraphics[scale=0.44]{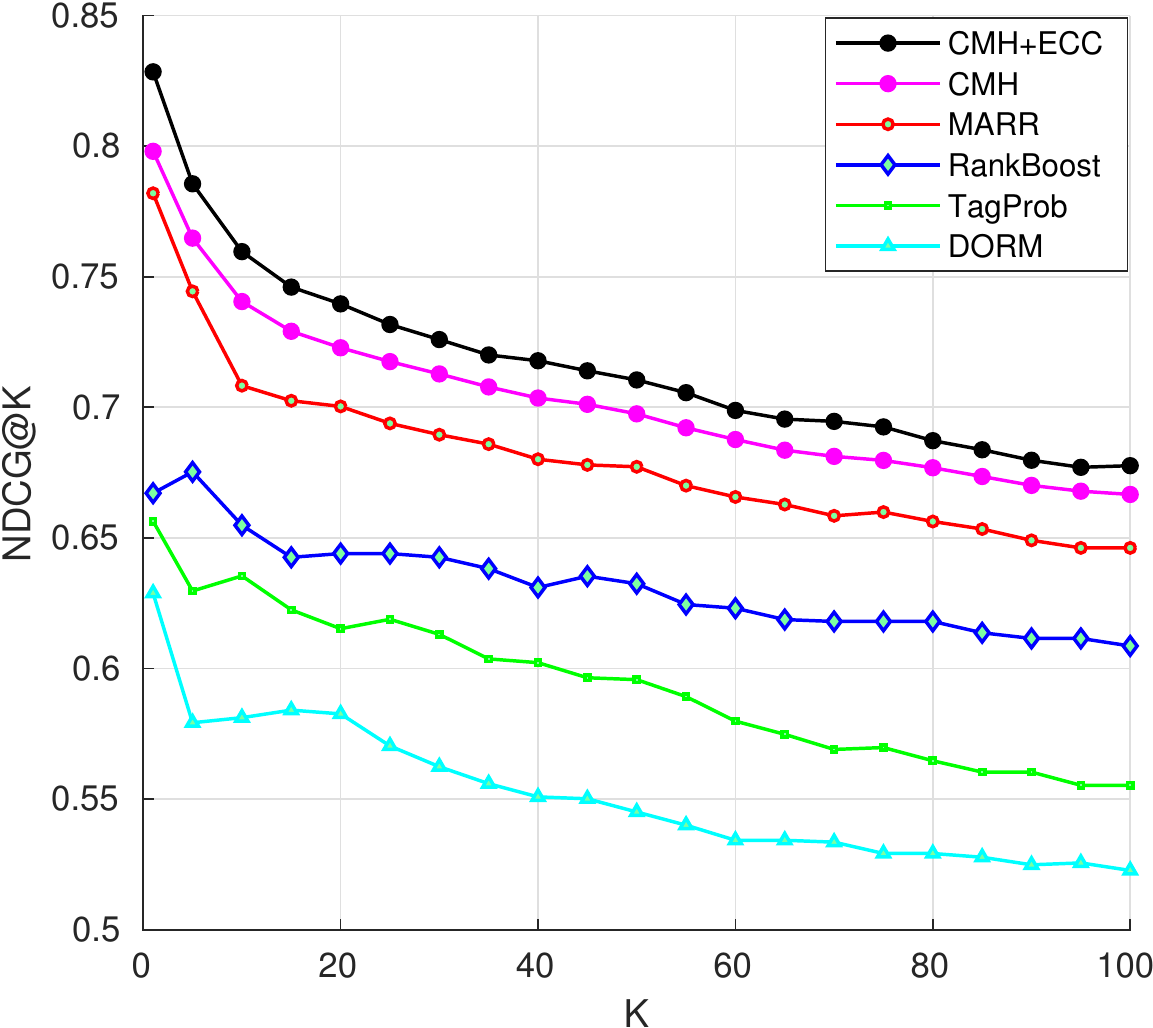}}
\subfigure[Double Attribute Queries]{\includegraphics[scale=0.4]{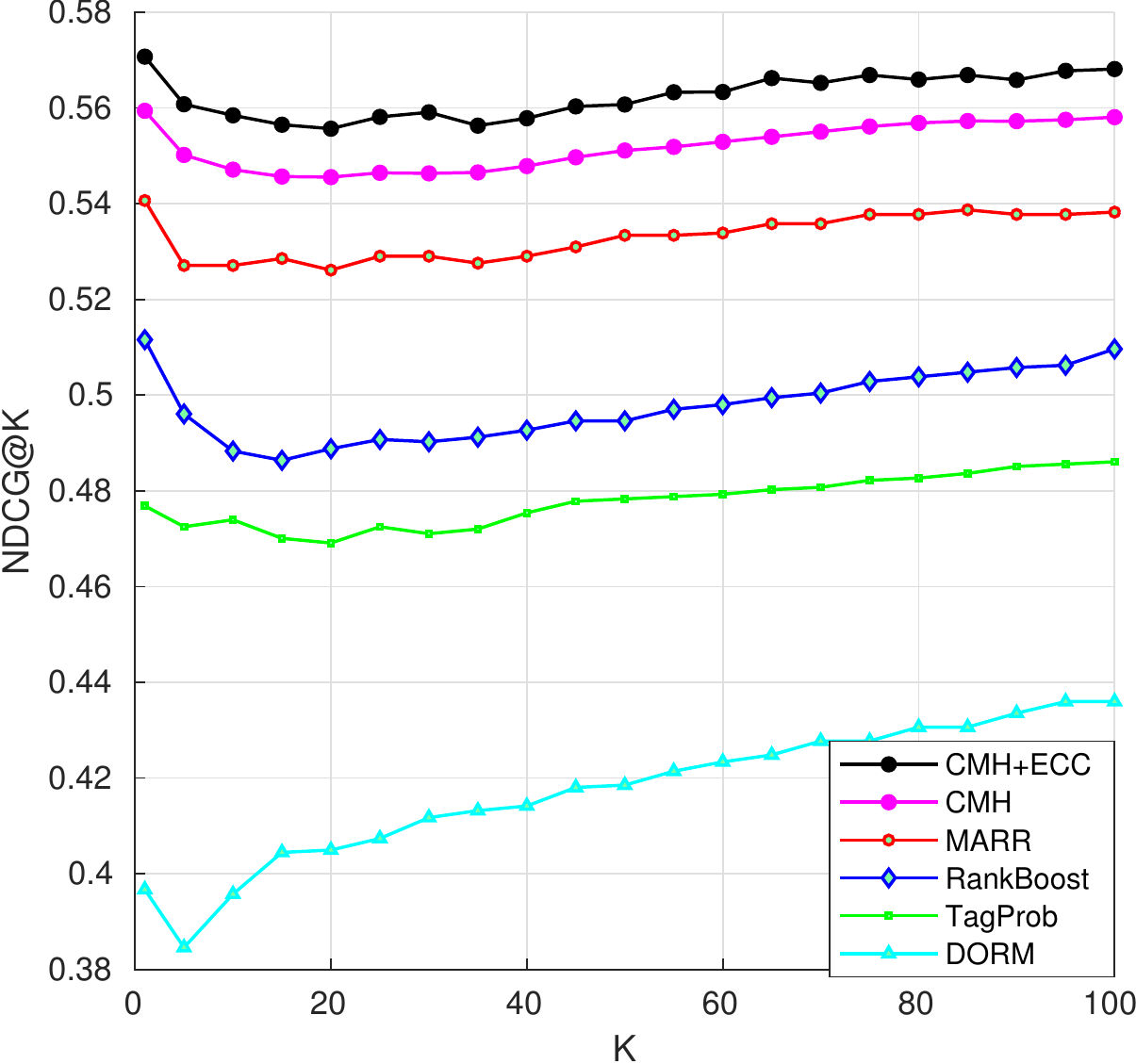}}
\subfigure[Triple Attribute Queries]{\includegraphics[scale=0.4]{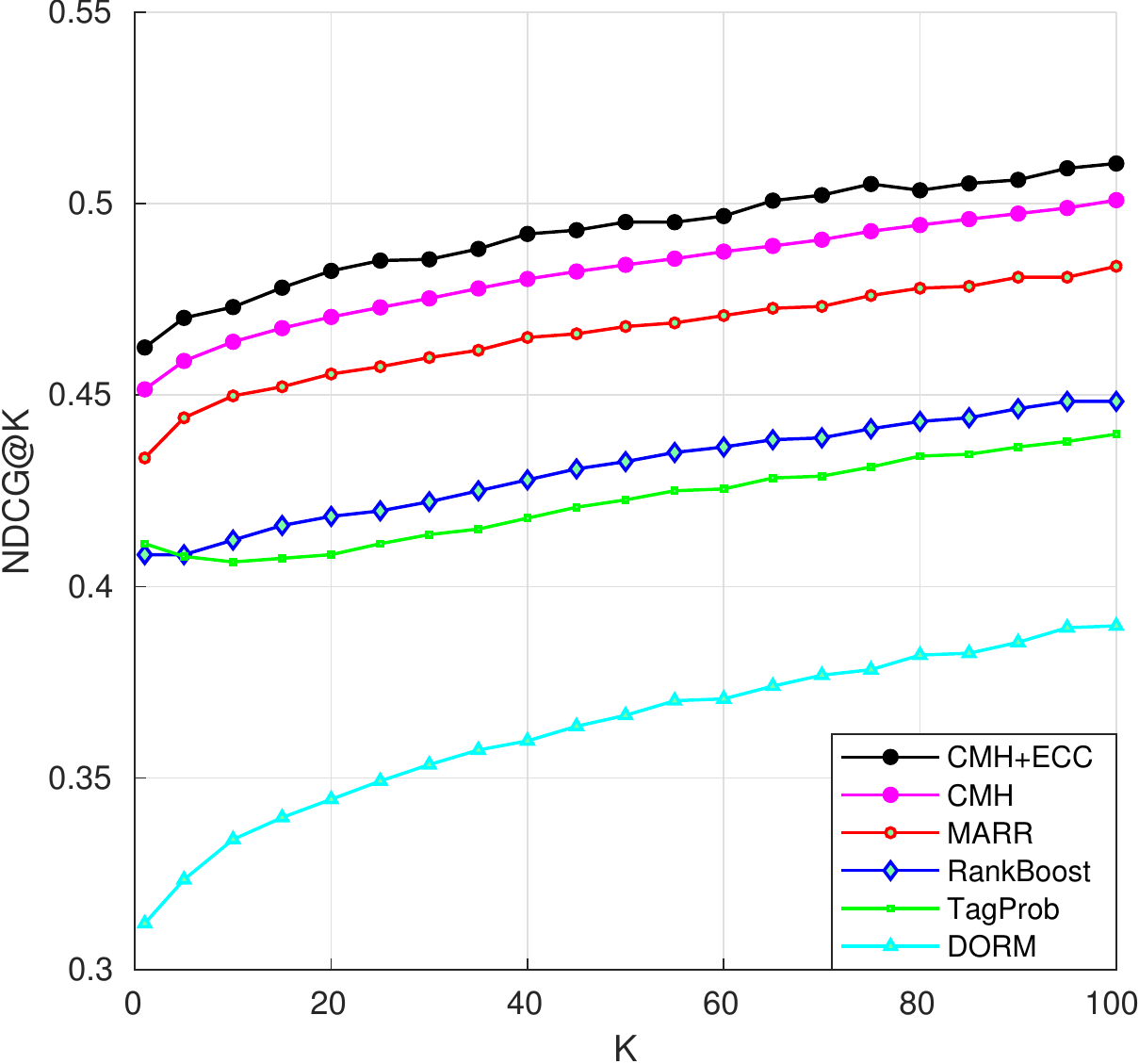}}
\vspace{-0.25cm}
\caption{ Ranking performance on the LFW dataset. } \label{fig:lfw}
\vspace{-0.35cm}
\end{figure*}
\begin{figure*}[t]
\centering
\subfigure[Single Attribute Queries]{\includegraphics[scale=0.4]{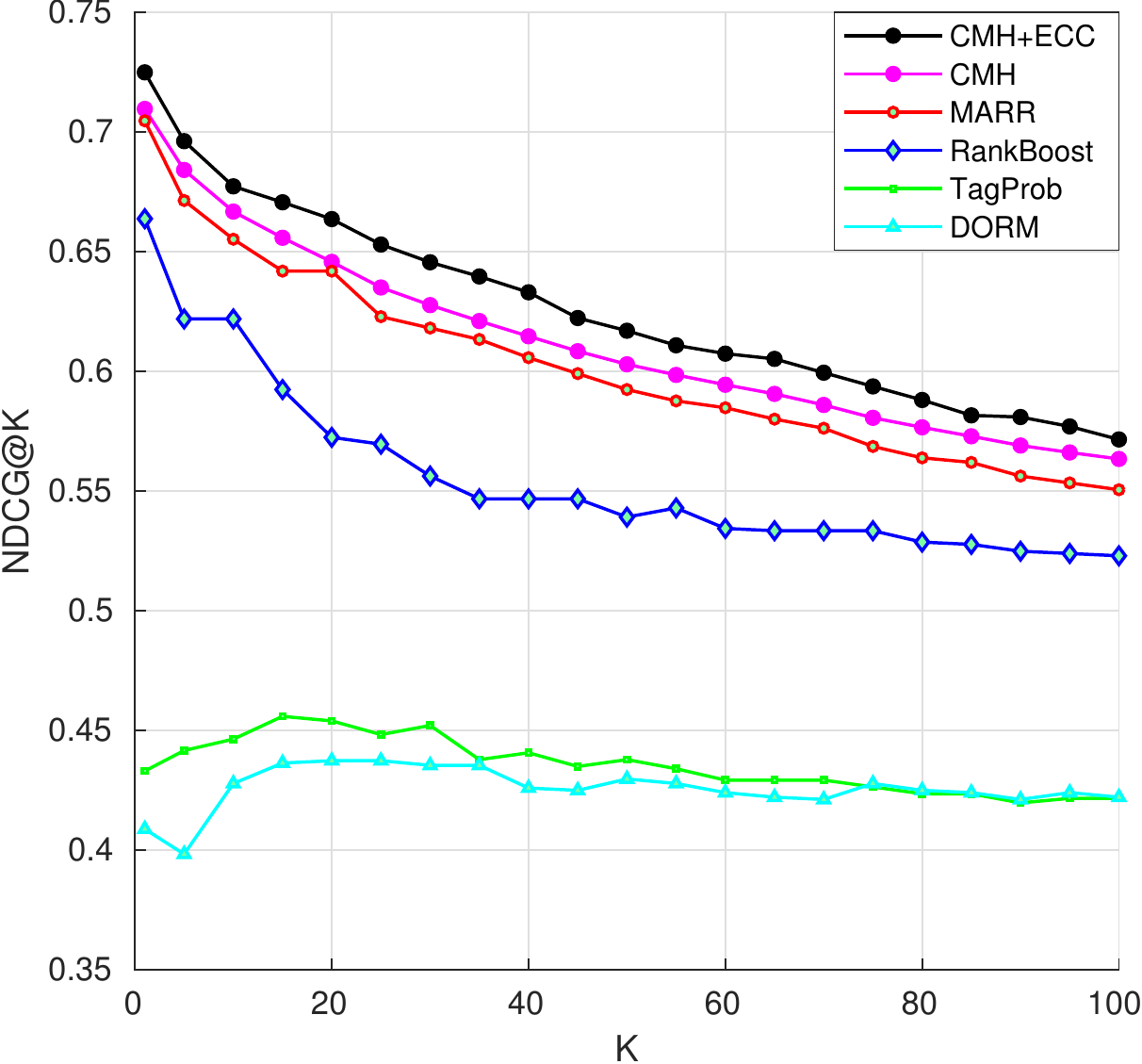}}
\subfigure[Double Attribute Queries]{\includegraphics[scale=0.4]{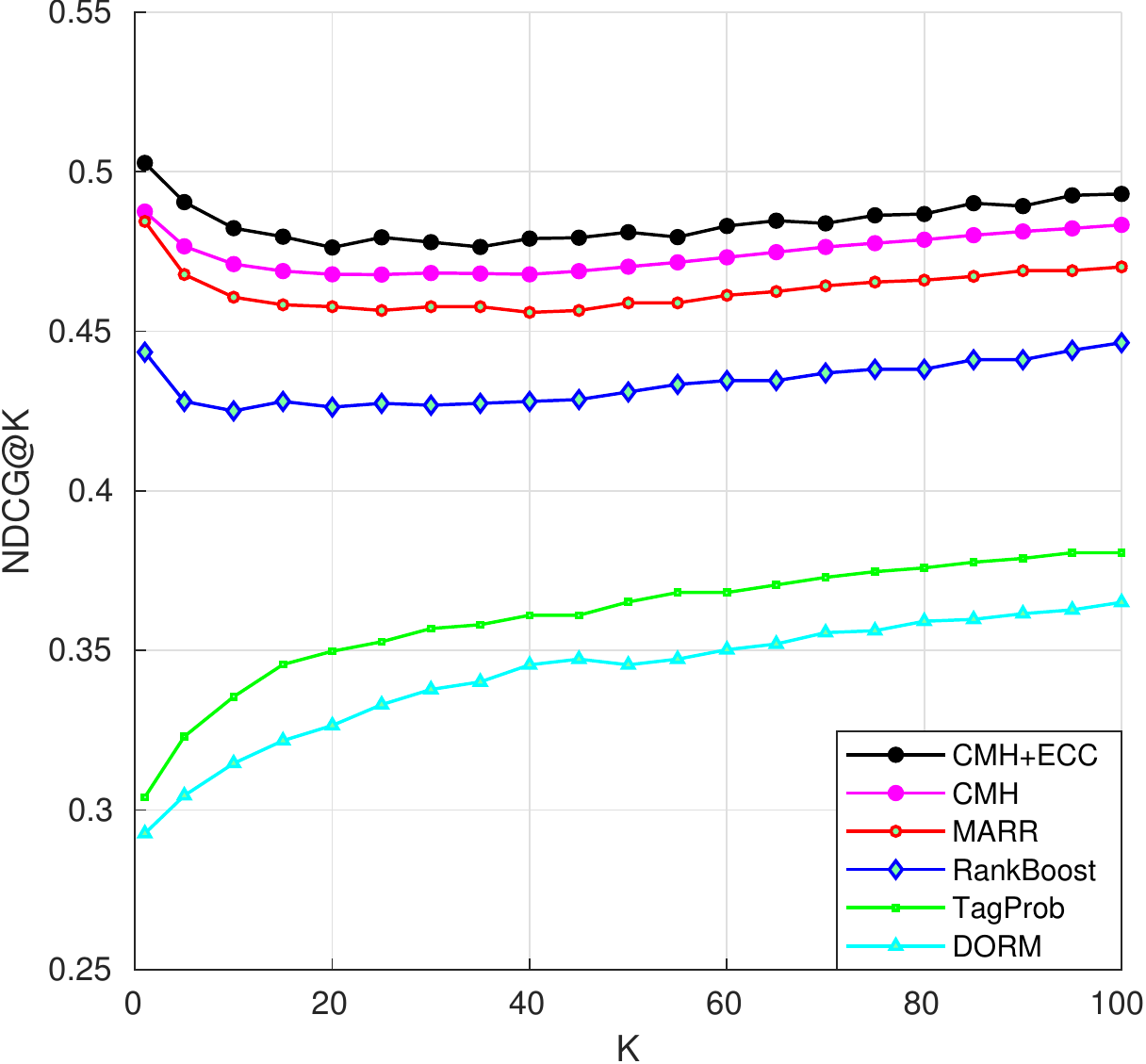}}
\subfigure[Triple Attribute Queries]{\includegraphics[scale=0.4]{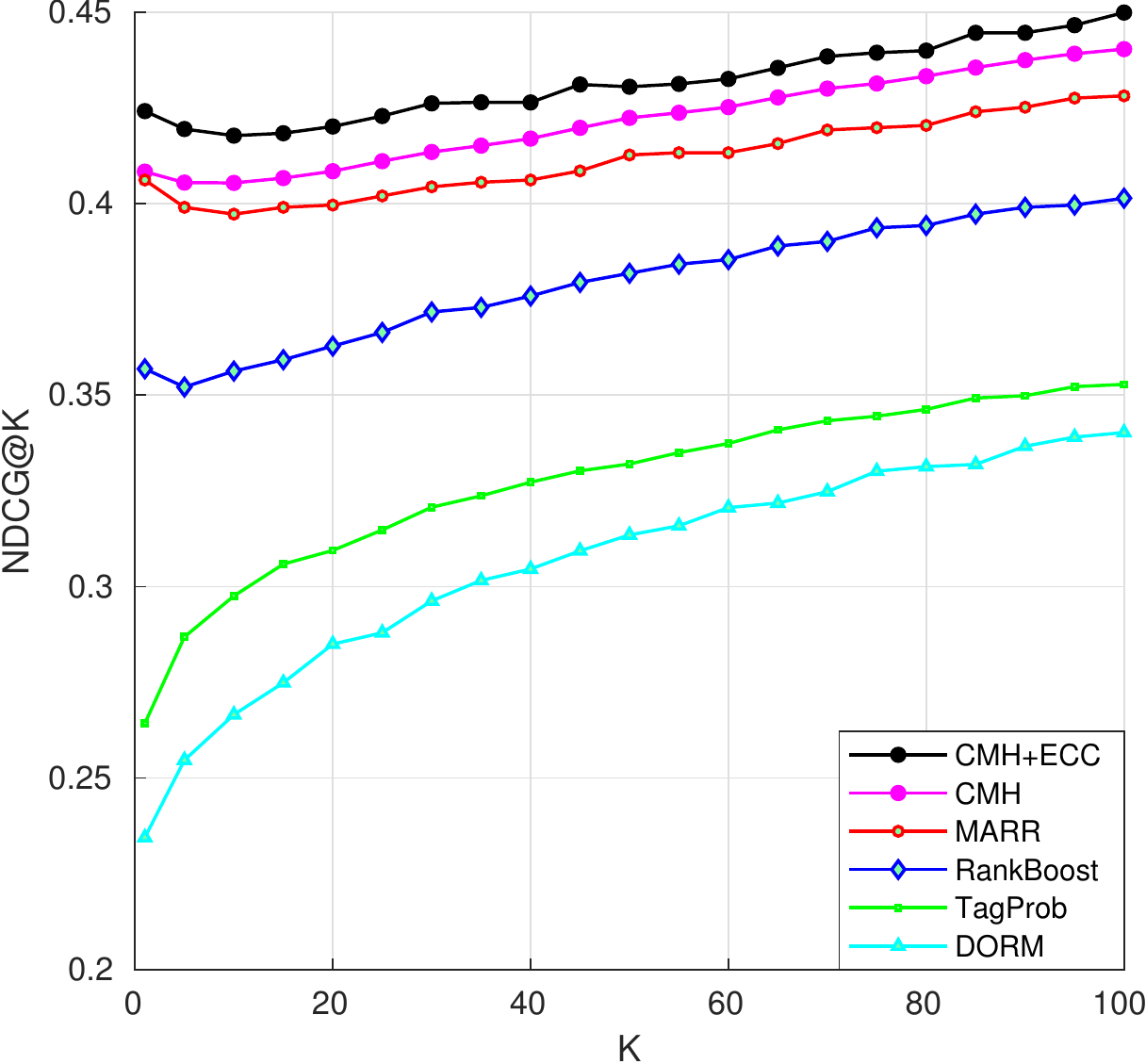}}
\vspace{-0.25cm}
\caption{ Ranking performance on the FaceTracer dataset. } \label{fig:facetracer}
\vspace{-0.35cm}
\end{figure*}

 Let $ f(\textbf{w}_x, \textbf{x}\textsubscript{i}) \in \mathbb{R}^d $ and $ g(\textbf{w}_y, \textbf{y}\textsubscript{j}) $ represent the learned CNN features for image modality $\textbf{x}\textsubscript{i}$  and MLP features for attribute modality $\textbf{y}\textsubscript{j}$, respectively. $\textbf{w}_x$ and $\textbf{w}_y$ are the CNN network weights and the MLP network weights, respectively. We define the total objective function for CMH as follows:
\begin{equation}
\begin{aligned}
    \underset{\textbf{C}_{x,y}, \textbf{w}_x, \textbf{w}_y }{\text{\ \ \ \ \ min\ \ \ \ \ }}   & \mathcal{J}=  \sum_{i=1}^n \sum_{j=1}^n   \underbrace  {\ell_c( p(\textbf{F}_{*i},\textbf{G}_{*j}),S_{ij})}_ \text{distance- based logistic loss}  + \\ &   \alpha \underbrace{(|| \textbf{F}-\textbf{C}_x||_F^2+||\textbf{G}-\textbf{C}_y||_F^2)}_\text{quantization loss} 
    +  \\  & \beta \underbrace{(|| \textbf{F}\textbf{1} ||_F^2+||\textbf{G}\textbf{1} ||_F^2)}_\text{entropy maximization}  \text{\ \ \ s.t. \ } \textbf{C}_{x,y} \in \{+1, -1 \}^{d \times n},
\end{aligned} \tag{1} \label{eq:1}   
\end{equation}
where $\textbf{F} \in  \mathbb{R}^{d \times n}$ is the image feature matrix and $\textbf{G} \in  \mathbb{R}^{d \times n}$ is the facial attribute feature matrix constructed by placing column-wise CNN and MLP features of training samples respectively. $\textbf{F}_{*i}= f(\textbf{w}_x, \textbf{x}\textsubscript{i})$ is the CNN feature corresponding to sample $\textbf{x}\textsubscript{i}$ and $\textbf{G}_{*j}= f(\textbf{w}_y, \textbf{y}\textsubscript{j})$ is the MLP feature corresponding to sample $\textbf{y}\textsubscript{j}$. $\textbf{C}_{x}$ and $\textbf{C}_{y}$ are the binary hash code matrices for image and attribute modalities, respectively. Notation \textbf{1} represents a vector with all its elements set to 1.

The first term in the objective function is distance-based logistic loss. This loss causes modalities referring to the same sample to attract one another and modalities to repel if they refer to two different samples. The distance based logistic-loss is derived from distance-based logistic probability, which is given by 
$p(\textbf{F}_{*i},\textbf{G}_{*j})=\frac{1+exp(-m)}{1+exp(||\textbf{F}_{*i}-\textbf{G}_{*j}||-m)}$ and represents the probability of the match between the image modality feature vector $\textbf{F}_{*i}$ and attribute modality feature vector $\textbf{G}_{*j}$, given their squared distance. The margin parameter $m$ determines the extent to which matched or non-matched samples are attracted or repelled, respectively. Then we apply the cross entropy loss similar to the classification case for deriving the final distance-based logistic loss : $\ell_c(p,s)=-s log (p) +(s-1) log(1-p)$. The second term in the objective function helps us to preserve the cross-modal similarity in the binary domain using hash codes $\textbf{C}_x$ and $\textbf{C}_y$, where $\textbf{C}_x = \mathsf{sign}(\textbf{F}) $ and $\textbf{C}_y = \mathsf{sign}(\textbf{G})$. The third term in the objective function attempts to maximize the entropy on the bits of the  hash code by making each bit of the hash code be balanced on all the training points. Precisely, the number of +1 and −1 for each bit on all the training samples should be almost the same. $\alpha$ and $\beta$ are tuning parameters that we set to 1. 

\vspace{-0.30cm}

\subsection {Error correcting code module (ECC)}

The intermediate hash codes generated by the CMH module can be used for a retrieval process. However, after gaining experience with CMH, we have concluded that there is an opportunity for improvement and further reducing the Hamming distance for different modalities of the same subject. On further research and inspired by \cite{talreja_multibiometric_2017}, we identified error correcting codes to be a promising solution for reducing the Hamming distance for different modalities of the same subject. 

We assume that the intermediate hash code generated by the CMH module is a binary vector that is within a certain distance from a codeword of an error-correcting code. By passing the intermediate hash code through an appropriate FEC decoder, the closest codeword is found and this closest codeword is used as a final hash code for the retrieval process. The main component of the ECC module is the forward error correcting (FEC) decoder. Due to their minimum-distance separable (MDS) property and widely available hardware, we have adopted Reed Solomon (RS) codes as our form of coding for FEC decoder.

The RS codes use symbols of length $m$ bits. Using a symbol size of $m$ bits, the length of the RS codeword is given by $N= 2^{m-1}$ in symbols, which corresponds to $n=mN$ in bits. However, we have utilized shortened RS codes for designing the FEC decoder of the ECC module. A shortened RS code is one in which the input to the decoder given as $N_1$ is less than the actual codeword length $N=2^{m-1}$. For our decoder, we have used shortened RS code with $m=8$ and $N=255$ symbols and the input to the decoder $N_1$ equal to $32$ symbols which is equal to $256$ bits. The intermediate hash code, which is used as the input to the FEC decoder $N_1$ is taken as 256 bits for all the experiments.  The codewords generated after decoding correspond to the final hash codes which are used for retrieval process using Hamming distance.

\vspace{-0.35cm}

\section{Experimental  Results}


\textbf{Datasets}: FaceTracer \cite{kumar2008facetracer} and LFW \cite{huang2007labeled} datasets have been used to evaluate our proposed framework. LFW is a popular dataset of more than 13,000 images of faces collected from the internet for face recognition as well as attribute classification. The FaceTracer dataset is a large collection of $15,000$ real-world face images, collected from the internet.

\textbf{Evaluation Results}: We follow the experimental protocol used in Multi Attribute Retrieval and Ranking (MARR) \cite{2011_siddiquie_image_ranking}. We use normalized discounted cumulative gain (NDCG) as our evaluation metric to compare CMH-ECC performance with other methods. NDCG is a standard single-number measure of ranking quality that allows non-binary relevance judgments, while most traditional ranking measures only allow binary relevance (relevant or not relevant).  NDCG is defined as  
$ NDCG @ k=\frac{1}{Z} \sum_{i=1}^k\frac{2^{rel(i)}-1}{log(i+1)} $, where rel(i) is the relevance of the $i^{th}$ ranked image and Z is a normalization constant to ensure that the correct ranking results in an NDCG score of 1.


Fig. \ref{fig:results} indicates the qualitative result of CMH-ECC approach for the given facial attributes. We compare the ranking quality using NDCG scores of the proposed CMH-ECC with four state of the art retrieval methods including MARR \cite{2011_siddiquie_image_ranking}, rankBoost \cite{freund_2003_efficient}, Direct Optimization of Ranking Measures (DORM) \cite{quoc_2007_dorm}, TagProp \cite{guillaumin_2009_tagprop}. In addition, we have also compared our results of CMH-ECC framework with the results of using only CMH module without FEC, which implies using intermediate hash codes as our final hash codes. Fig. \ref{fig:lfw} and Fig. \ref{fig:facetracer} plots the NDCG scores, as a function of the ranking truncation level \textit{k}, using different number of attribute queries for the LFW and FaceTracer dataset, respectively. We can observe that CMH-ECC generally outperforms the comparison methods for both datasets using all the three types of queries. In particular, compared to the state of the art method MARR, we achieve approximately an increase of $4.0\%$, $3.5\%$ and $3.0\%$ in NDCG values for single, double and triple attribute queries, respectively. The retrieval efficiency using intermediate hash codes generated by our CMH module also outperforms MARR. Notice that NDCG values for the FaceTracer dataset for all the methods are relatively lower when compared to the LFW dataset. This is due to the difference in the distributions of the two datasets.

\vspace{-0.30cm}
\section{Conclusion}
\vspace{-0.25cm}
In this paper, we proposed a facial retrieval algorithm using deep hashing network and forward error correcting decoder to retrieve relevant facial images from the database using a given attribute query. This is the first time where error correcting codes have been combined with deep cross modal hashing for image retrieval. The experimental results on two popular public datasets show that our method outperforms the current face image retrieval approaches in the literature. 


\vfill\pagebreak

\bibliographystyle{IEEEbib}
\bibliography{Eadcmh}

\end{document}